\documentclass[letterpaper]{article} 
\usepackage{aaai24}  
\usepackage{times}  
\usepackage{helvet}  
\usepackage{courier}  
\usepackage[hyphens]{url}  
\usepackage{graphicx} 
\usepackage{natbib}  
\usepackage{caption} 
\usepackage{algorithm}
\usepackage{algorithmic}

\usepackage{booktabs}
\usepackage{xcolor}
\usepackage{amsmath}

\usepackage{newfloat}
\usepackage{listings}
\DeclareCaptionStyle{ruled}{labelfont=normalfont,labelsep=colon,strut=off} 
\floatstyle{ruled}
\newfloat{listing}{tb}{lst}{}
\floatname{listing}{Listing}
\title{Benchmarking Large Language Models in Retrieval-Augmented Generation}
\author{
    Jiawei Chen\textsuperscript{\rm 1,3},
    Hongyu Lin\textsuperscript{\rm 1,}\thanks{~ Corresponding authors.},
    Xianpei Han\textsuperscript{\rm 1,2,}\footnotemark[1],
    Le Sun\textsuperscript{\rm 1,2}
}
\affiliations{
    \textsuperscript{\rm 1}Chinese Information Processing Laboratory ~
    \textsuperscript{\rm 2}State Key Laboratory of Computer Science\\
    Institute of Software, Chinese Academy of Sciences, Beijing, China \\
    \textsuperscript{\rm 3}University of Chinese Academy of Sciences, Beijing, China

    \{jiawei2020,hongyu,xianpei,sunle\}@iscas.ac.cn
}

\begin{document}

\maketitle

\begin{abstract}
Retrieval-Augmented Generation (RAG) is a promising approach for mitigating the hallucination of large language models (LLMs). However, existing research lacks rigorous evaluation of the impact of retrieval-augmented generation on different large language models, which make it challenging to identify the potential bottlenecks in the capabilities of RAG for different LLMs. In this paper, we systematically investigate the impact of Retrieval-Augmented Generation on large language models. We analyze the performance of different large language models in 4 fundamental abilities required for RAG, including noise robustness, negative rejection, information integration, and counterfactual robustness. To this end, we establish Retrieval-Augmented Generation Benchmark (RGB), a new corpus for RAG evaluation in both English and Chinese. RGB divides the instances within the benchmark into 4 separate testbeds based on the aforementioned fundamental abilities required to resolve the case. Then we evaluate 6 representative LLMs on RGB to diagnose the challenges of current LLMs when applying RAG. Evaluation reveals that while LLMs exhibit a certain degree of noise robustness, they still struggle significantly in terms of negative rejection, information integration, and dealing with false information. The aforementioned assessment outcomes indicate that there is still a considerable journey ahead to effectively apply RAG to LLMs.
\end{abstract}

\section{Introduction}
Recently, there have been impressive advancements in large language models (LLMs) like ChatGPT~\citep{chatgpt} and ChatGLM~\citep{chatglm}. Although these models have shown remarkable general abilities~\citep{bang2023multitask,guo2023close}, they still suffer severely from challenges including factual  hallucination~\citep{cao-etal-2020-factual,raunak-etal-2021-curious,10.1145/3571730}, knowledge out-dating~\citep{he2022rethinking}, and the lack of domain-specific expertise~\citep{li2023chatgpt,shen2023chatgpt}.

Incorporating external knowledge via information retrieval, i.e., Retrieval-Augmented 
Generation (RAG), has been regarded as a promising way to resolve the above challenges. ~\citep{10.5555/3524938.3525306,10.5555/3495724.3496517,borgeaud2022improving,izacard2022atlas}. With the help of external knowledge, LLMs can generate more accurate and reliable responses. The most common method is to use a search engine as a retriever such as New Bing. Due to the vast amount of information available on the Internet, using a search engine can provide more real-time information.

\begin{figure}[t!]
\centering 
\includegraphics[width=0.45\textwidth]{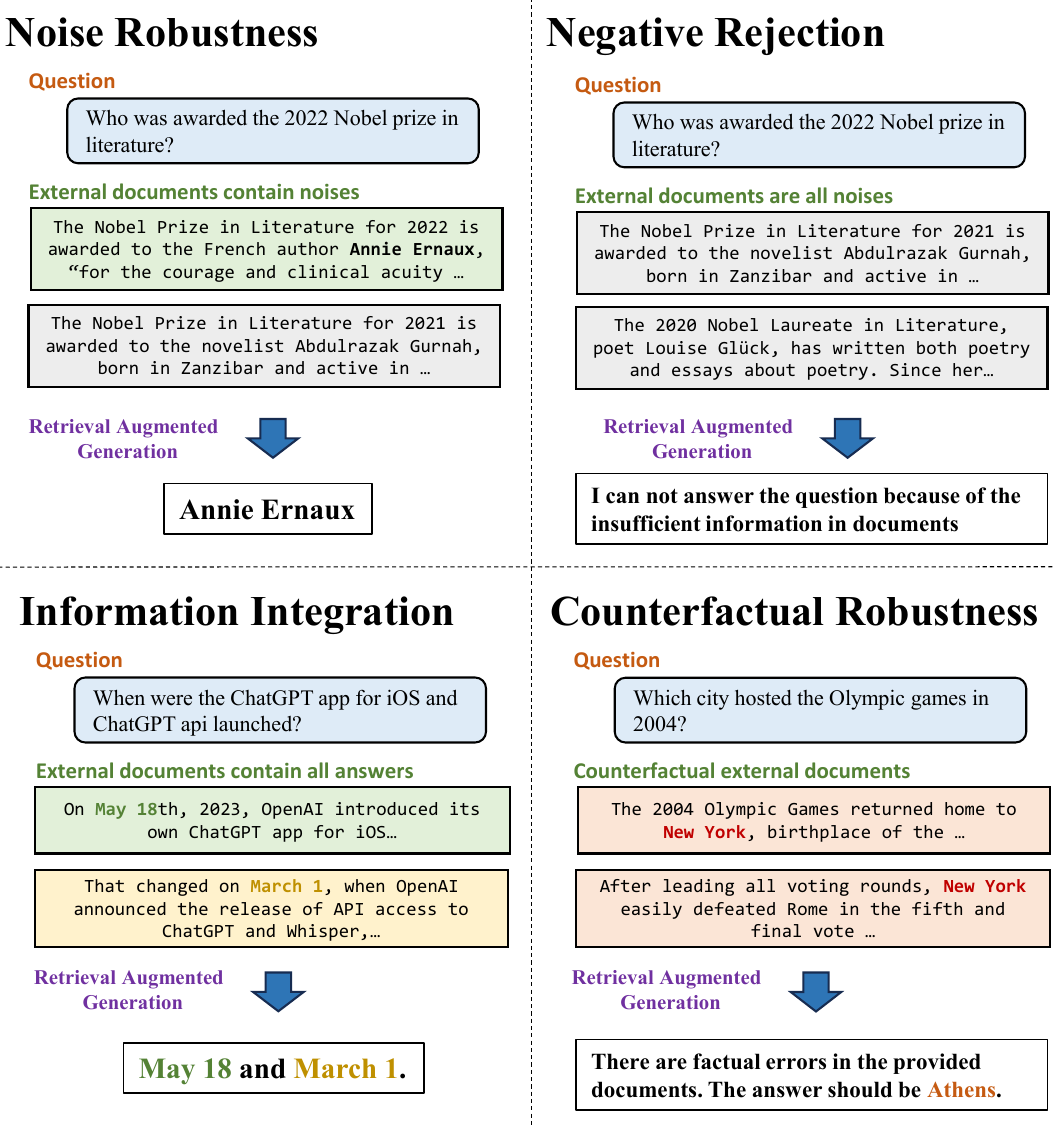}
\caption{Illustration of 4 kinds of abilities required for retrieval-augmented generation of LLMs.}
\label{fig:exam}
\end{figure}

However, Retrieval-Augmented Generation brings not only positive effects to LLMs~\citep{liu2023evaluating,maynez-etal-2020-faithfulness}. On one hand, there is a significant amount of noise information even fake news in the content available on the Internet, which poses challenges for search engines in accurately retrieving desirable knowledge. On the other hand, LLMs suffer from unreliable generation challenge. LLMs can be misled by incorrect information contained in the context~\citep{bian2023drop} and also suffer from hallucination during the generation~\citep{adlakha2023evaluating}, resulting in generating content that goes beyond external information. These challenges result in LLMs being unable to consistently generate reliable and accurate responses. Unfortunately, currently there lacks of comprehensive understanding on how these factors can influence RAG, and how could each model survives from these drawbacks and improvement their performance via information retrieval. As a result,  there is a pressing need for a comprehensive evaluation of LLMs on their ability to effectively utilize retrieved information, as well as their ability to withstand the various drawbacks present in information retrieval.

To this end, this paper conducts a comprehensive evaluation of RAG for current LLMs. Specifically, we create a new Retrieval-Augmented Generation Benchmark, namely RGB, in both English and Chinese. In order to ensure that the internal knowledge of LLMs does not introduce bias into the evaluation results, RGB chooses to aggregate the latest news information and constructs queries based on the news information. Then, based on these queries, we use Search API to fetch relevant documents and select most relevant snippets from the content as external retrieved documents. Finally, based on different compositions of query and document-set pairs, we expand the corpus and divided it into 4 testbeds to evaluate the following basic abilities of LLMs according to the common challenges in RAG, as shown in Figure~\ref{fig:exam}:
\begin{itemize}
    \item \textbf{Noise Robustness}, which means a LLM can extract useful information from noisy documents. In this paper, we define noisy documents as those that are relevant to the question but do not contain any information of the answer. For the instance in Figure~\ref{fig:exam}, the noisy documents related to the question ``Who was awarded the 2022 Nobel Prize in Literature'' include reports about the 2021 Nobel Prize in Literature. To this end, the testbed for noise robustness contains instances whose external documents contain a certain number of noisy documents based on the desired noise ratio. 
    \item \textbf{Negative Rejection}, which means that a LLM should reject to answer the question when the required knowledge is not present in any retrieved document. The testbed for negative rejection contains instances whose external documents are only with noisy documents. LLMs are expected to indicate ``insufficient information'' or other rejection signals.
    \item \textbf{Information Integration}, which evaluates whether LLMs can answer complex questions that require integrating information from multiple documents. For the instance in Figure~\ref{fig:exam}, for the question ``When were the ChatGPT app for iOS and ChatGPT api launched?'', LLMs are expected to provide information of the launch dates for both the ChatGPT iOS app and ChatGPT API. The testbed for information integration contains instances that can only be answered using multiple documents.
    \item \textbf{Counterfactual Robustness}, which evaluates whether LLMs can identify risks of known factual errors in the retrieved documents when the LLMs are given warnings about potential risks in the retrieved information through instruction. The testbed for counterfactual robustness includes instances that can be answered directly by the LLMs, but the external documents contain factual errors.
\end{itemize}

Based on RGB, we conduct evaluation on 6 state-of-the-art large language models including ChatGPT~\citep{chatgpt}, ChatGLM-6B~\citep{chatglm}, ChatGLM2-6B~\citep{chatglm2}, Vicuna-7b~\citep{vicuna2023}, Qwen-7B-Chat~\citep{qwen}, BELLE-7B~\citep{BELLE}. We found that even though RAG can improve the response accuracy of LLMs, they still suffer from the above-mentioned challenges significantly. Specifically, we found that even though LLMs demonstrate some level of noise robustness, they tend to confuse similar information and frequently generate inaccurate answers when relevant information exists. For example, when faced with a question about the 2022 Nobel Prize in Literature, if there are noisy documents about the 2021 Nobel Prize in Literature in external documents, LLMs may become confused and provide inaccurate answers. Besides, LLMs frequently fail to reject answering and generate incorrect answers when none of the external documents contain relevant information. Furthermore, LLMs lack the ability to summarize from multiple documents, and therefore if multiple documents are needed to answer a question, LLMs often fail to provide accurate answer. Finally, we found that even when the LLMs contain the required knowledge and are given warnings about potential risks in the retrieved information through instruction, they still tend to trust and prioritize the retrieved information over their own existing knowledge. The experimental results mentioned above highlight the need for further resolution of important issues in the existing RAG method. Therefore, it is crucial to exercise caution and carefully design its usage.

Generally speaking, the contributions of this paper are\footnote{Our code\&data: \url{https://github.com/chen700564/RGB}.}:
\begin{itemize}
    \item We proposed to evaluate four capabilities for retrieval-augmented generation of LLMs and created the Retrieval-Augmented Generation Benchmark in both English and Chinese. To best of our knowledge, it is the first benchmark designed to assess these four capabilities for retrieval-augmented generation of LLMs.
    \item We evaluated the existing LLMs using RGB and found the limitations of them in the four different abilities.
    \item We analyzed the responses of LLMs in RGB and identified their current shortcomings as well as suggested directions for improvement.
\end{itemize}

\section{Related work}
\paragraph{Retrieval-augmented models} 

The knowledge stored in large language models is commonly out-of-date~\citep{he2022rethinking} and they also sometimes generate hallucination~\citep{cao-etal-2020-factual,raunak-etal-2021-curious,10.1145/3571730} i.e., they may generate irrelevant or factually incorrect contents. By using external knowledge as guidance, retrieval-augmented models can generate more accurate and reliable responses~\citep{10.5555/3524938.3525306,10.5555/3495724.3496517,borgeaud2022improving,izacard2022atlas,shi2023replug,ren2023investigating}. Retrieval-augmented models have achieved remarkable results in various tasks such as open-domain QA~\citep{izacard-grave-2021-leveraging,trivedi-etal-2023-interleaving,li-etal-2023-large}, dialogue~\citep{cai-etal-2019-skeleton,cai-etal-2019-retrieval,peng2023check}, domain-specific question answering~\citep{cui2023chatlaw} and code generation~\citep{zhou2023docprompting}. Recently, with the development of large models, a series of retrieval-enhanced tools and products have gained widespread attention, such as ChatGPT retrieval plugin, Langchain, New Bing, etc. However, in real-world scenarios, the retrieved text inevitably contains noise. Therefore, in this paper we conducted a systematic evaluation and analysis of retrieval-augmented generation in LLMs.

\begin{figure}[t!]
\centering 
\includegraphics[width=0.4\textwidth]{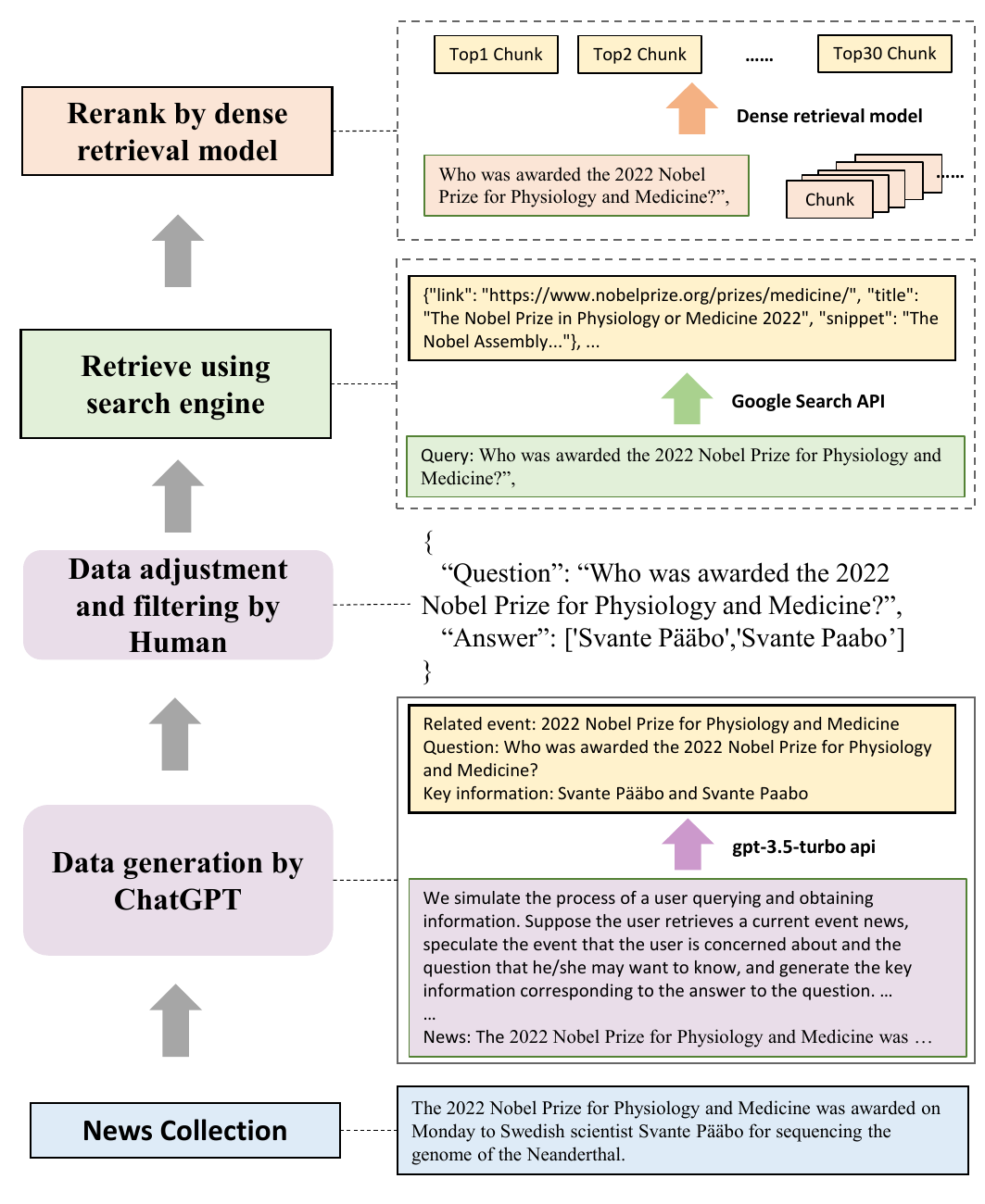}
\caption{The process of data generation. Firstly, we use models to extract (event, question, answer) from news articles. Next, we utilize search engines to retrieve relevant web pages. Finally, a dense retrieval model is employed to re-rank the content of these web pages.}
\label{fig:data}
\end{figure}

\paragraph{Evaluation of LLMs}
Evaluating LLMs has received significant attention due to their remarkable general capability~\citep{chang2023survey}. It enables us to gain a deeper understanding of the specific abilities and limitations of LLMs, while also providing valuable guidance for future research. In the past, benchmarks such as GLUE~\citep{wang2018glue} and SuperCLUE~\citep{10.5555/3454287.3454581} primarily focused on evaluating NLP tasks, particularly in natural language understanding. However, these evaluations often fail to fully capture the capabilities of LLMs. MMLU~\citep{hendrycks2021measuring} was then proposed to measure the knowledge acquired by language models when pre-training. Recently, with the development of LLMs, a series of general evaluation benchmarks have emerged, such as AGIEval~\citep{zhong2023agieval}, C-Eval~\citep{huang2023ceval}, AlpacaEval~\citep{alpaca_eval}, OpenLLM Leaderboard~\citep{open-llm-leaderboard}, etc. In addition to general abilities, there are also specific benchmarks that focus on evaluating the capabilities of models. For example, CValues~\citep{xu2023cvalues} focuses on the safety and responsibility of LLMs, M3Exam~\citep{zhang2023m3exam} focuses on human exam and ToolBench~\citep{qin2023toolllm} evaluates how well LLMs use external tools. Recently, \citet{adlakha2023evaluating} evaluate the RAG of LLMs in exist QA dataset. Different from their work, we focus on 4 required abilities of RAG and create Retrieval-Augmented Generation Benchmark to evaluate the LLMs.
\section{Retrieval-Augmented Generation Benchmark}
In this section, we first introduce the specific retrieval-augmented generation abilities we aim to evaluate. Next, we outline the process of constructing the RAG benchmark for evaluation. Lastly, we present the evaluation metrics.

\subsection{Required abilities of RAG}
External knowledge is the key to resolving the problems of LLMs such as hallucination and outdated knowledge, which can make LLMs generate more accurate and reliable responses through retrieval-augmented generation (RAG). However, LLMs cannot always response as expected with RAG. For one thing, there are numerous irrelevant documents and false information on the Internet. Incorporating these external documents into LLMs could have a detrimental effect. For anthoer, LLMs suffer from the unreliable generation challenge. The generation of LLMs is often unpredictable, and we cannot guarantee that they will utilize the useful information entailed in the external documents. Additionally, LLMs can easily be misled by incorrect information in the document. To this end, we build Retrieval-Augmented Generation Benchmark (RGB) to evaluate the retrieval-augmented generation of LLMs, and we concern about 4 specific abilities:

\textbf{Noise Robustness} is the robustness of LLMs in noisy documents. As retrievers are not perfect, the external knowledge they retrieve often contains a significant amount of noise, i.e., documents which are relevant to the question but do not contain any information about the answer. To effectively answer user questions, LLMs must be able to extract the necessary information from documents despite there are noisy documents.

\textbf{Negative Rejection} is a measure of whether LLMs can decline to answer a question when none of the contexts provide useful information. In real-world situations, the search engine often fails to retrieve documents containing the answers. In these cases, it is important for the model to have the capability to reject recognition and avoid generating misleading content.

\textbf{Information Integration} is a capacity to integrate answers from multiple documents. In many cases, the answer to a question may be contained in multiple documents. For example, for the question \emph{``Who are the champions of the U.S. Open 2022 men's and women's singles?''}, the two champions may be mentioned in different documents. In order to provide better answers to complex questions, it is necessary for LLMs to have the ability to integrate information.

\textbf{Counterfactual Robustness} refers to a capacity to handle errors in external knowledge. In the real world, there is an abundance of false information on the internet. Please note that we only evaluate the situation that LLMs are given warnings about potential risks in the retrieved information through instruction.

In real-world scenarios, it is not possible to obtain perfect documents with all the necessary external knowledge. Therefore, evaluating these four abilities of the model becomes essential in order to measure the RAG of LLMs.

\subsection{Data construction}
Inspired by previous benchmarks for LLMs, RGB utilizes a question-answering format for evaluation. We evaluate the LLMs by judging the retrieval-augmented responses of them to the questions. To simulate real-world scenarios, we construct question and answer data using actual news articles.  Due to the abundance of knowledge contained within the LLMs there is a potential for bias when measuring the first three abilities. To mitigate this, the instances of RGB are constructed by latest news articles. Additionally, we retrieve external documents from Internet through search engines. Finally, we expand the corpus and divided it into 4 testbeds to evaluate the above basic abilities of LLMs. The overall procedure of our data construction is illustrated in Figure~\ref{fig:data}. 

\textbf{QA instances generation.} We first collect latest news articles and use prompts to make ChatGPT generate events, questions, and answers for each articles. For example, as shown in the Figure~\ref{fig:data}, for a report about ``The 2022 Nobel Prize'', ChatGPT will generate corresponding event, question and provide key information for answering it. By generating events, the model is able to preliminarily filter out news articles that do not contain any events. After generation, we manually check the answer  and filter out data that is difficult to retrieve through search engines.

\textbf{Retrieve using search engine.} For each query, we use Google's API to fetch 10 relevant web pages and extract corresponding snippets of text from them. Simultaneously, we read these web pages and convert their textual content into text chunks with a maximum length of 300 tokens. Using an existing dense retrieval model~\footnote{Chinese: \url{https://huggingface.co/moka-ai/m3e-base}; English: \url{https://huggingface.co/sentence-transformers/all-mpnet-base-v2}.}, we select the top-30 text chunks that match the query most effectively. These retrieved text chunks, along with the snippets provided by the search API, will serve as our external documents. These documents will be divided into positive documents and negative documents based on whether they contain the answer. 

\textbf{Testbeds construction for each ability.}  We expand the corpus and divided it into 4 testbeds to evaluate the above basic abilities of LLMs. To evaluate the noise robustness, we sample varying numbers of negative documents according to the desired ratio of noises. For negative rejection, all the external documents are sampled from negative documents. For the information integration ability, we further construct data based on the above generated questions. This involves expanding or rewriting these questions so that their answers encompass multiple aspects. For example, the question ``Who won the MVP of Super Bowl 2023?'' can be rewrite as ``Who won the MVPs of Super Bowl 2022 and 2023?''. Consequently, answering such questions requires utilizing information from various documents. Different from the first three abilities, the data of counterfactual robustness is constructed solely based on the internal knowledge of the model. Based on the aforementioned generated questions mentioned above, we adopt ChatGPT to automatically generate its known knowledge. Specifically, we use prompts to allow the model to generate both questions and answers that are already known. For example, based on the question ``Who was awarded the 2022 Nobel Prize for Physiology and Medicine?'', the model will generate the known question  ``Who was awarded the 2021 Nobel Prize in Literature?'' and answer ``\emph{Abdulrazak Gurnah}''. We then manually verified the generated answers, and retrieve relevant documents as described above. In order to make documents contain factual errors, we manually modify the answers and replace the corresponding parts in the document.  

\begin{figure}[t!]
\centering 
\includegraphics[width=0.43\textwidth]{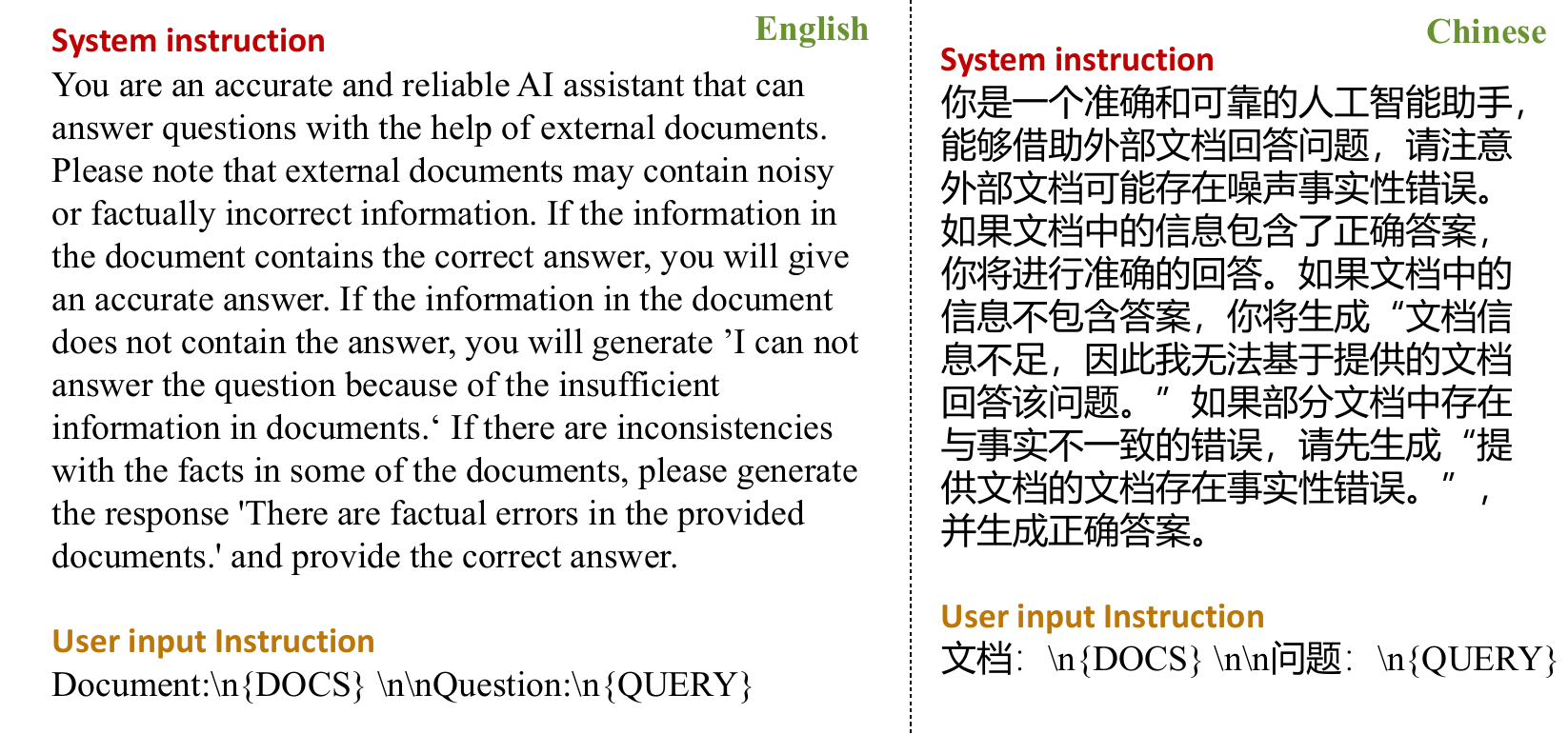}
\caption{The instructions used in our experiments, which include a system instruction followed by a user input instruction. The ``\{DOCS\}'' and ``\{QUERY\}'' will be replaced by the external documents and the question.}
\label{fig:io}
\end{figure}

\begin{table*}[]
\centering
\resizebox{0.8\textwidth}{!}{\begin{tabular}{@{}l|lllll|lllll@{}}
\toprule
\multicolumn{1}{c|}{}             & \multicolumn{5}{c|}{English}                                                                                                   & \multicolumn{5}{c}{Chinese}                                                                                                   \\ \midrule
\multicolumn{1}{c|}{Noise Ratio}   & \multicolumn{1}{c}{0} & \multicolumn{1}{c}{0.2} & \multicolumn{1}{c}{0.4} & \multicolumn{1}{c}{0.6} & \multicolumn{1}{c|}{0.8} & \multicolumn{1}{c}{0} & \multicolumn{1}{c}{0.2} & \multicolumn{1}{c}{0.4} & \multicolumn{1}{c}{0.6} & \multicolumn{1}{c}{0.8} \\ \midrule
ChatGPT~\citep{chatgpt}           & \textbf{96.33}        & \textbf{94.67}          & \textbf{94.00}          & \textbf{90.00}          & \textbf{76.00}           & \textbf{95.67}        & \textbf{94.67}          & \textbf{91.00}          & \textbf{87.67}          & \textbf{70.67}          \\
ChatGLM-6B~\citep{chatglm}        & 93.67                 & 90.67                   & 89.33                   & 84.67                   & 70.67                    & 94.33                 & 90.67                   & 89.00                   & 82.33                   & 69.00                   \\
ChatGLM2-6B~\citep{chatglm2}      & 91.33                 & 89.67                   & 83.00                   & 77.33                   & 57.33                    & 86.67                 & 82.33                   & 76.67                   & 72.33                   & 54.00                   \\
Vicuna-7B-v1.3~\citep{vicuna2023} & 87.67                 & 83.33                   & 86.00                   & 82.33                   & 60.33                    & 85.67                 & 82.67                   & 77.00                   & 69.33                   & 49.67                   \\
Qwen-7B-Chat~\citep{qwen}         & 94.33                 & 91.67                   & 91.00                   & 87.67                   & 73.67                    & 94.00                 & 92.33                   & 88.00                   & 84.33                   & 68.67                   \\
BELLE-7B-2M~\citep{BELLE}         & 83.33                 & 81.00                   & 79.00                   & 71.33                   & 64.67                    & 92.00                 & 88.67                   & 85.33                   & 78.33                   & 67.68                   \\ \bottomrule
\end{tabular}}
\caption{The experimental result of noise robustness measured by accuracy (\%) under different noise ratios. We can see that the increasing noise rate poses a challenge for RAG in LLMs.}
\label{tab:result1}
\end{table*}

\begin{table*}[]
\centering
\resizebox{0.95\textwidth}{!}{\begin{tabular}{@{}c|l|l|l@{}}
\toprule
            & \multicolumn{1}{c|}{\textbf{Long-distance information.}}                                                                                                                                                                                                                                                                                                              & \multicolumn{1}{c|}{\textbf{Evidence uncertainty.}}                                                                                                                                                                                                                                                                                      & \multicolumn{1}{c}{\textbf{Concept confusion.}}                                                                                                                                                                                                                                                                                                                                        \\ \midrule
\textbf{Question}  & Who did Iga Swiatek defeat to win the Qatar Open 2022?                                                                                                                                                                                                                                                                                                                & What is the name of Apple's headset?                                                                                                                                                                                                                                                                                                   & What was Tesla's revenue in Q1 2022?                                                                                                                                                                                                                                                                                                                                                                 \\ \midrule
\textbf{Answer}    & \textbf{Anett Kontaveit}                                                                                                                                                                                                                                                                                                                                              & \textbf{Vision Pro}                                                                                                                                                                                                                                                                                                                    & \textbf{18.76 billion}                                                                                                                                                                                                                                                                                                                                                                               \\ \midrule
\textbf{Documents} & \begin{tabular}[c]{@{}l@{}}\emph{Positive document}\\ In February, Swiatek entered into the \textcolor{blue}{Qatar Open} ...\\ In the final, she won ... \textbf{\textcolor{blue}{Anett Kontaveit}} ...\\ \\ \emph{Negative document}\\ This time, she defeated \textcolor{red}{Ons Jabeur} 6-2, 7-6(5) to win\\ the \textcolor{red}{2022 US Open}, ...\end{tabular} & \begin{tabular}[c]{@{}l@{}}\emph{Positive document}\\ Apple (AAPL.O) on Monday unveiled a costly\\ augmented-reality \textcolor{blue}{headset} called the \textbf{\textcolor{blue}{Vision Pro}} ...\\ \\ \emph{Negative document}\\ ... is what Gurman believes will be called \\ \textcolor{red}{Apple Reality Pro}. ...\end{tabular} & \begin{tabular}[c]{@{}l@{}}\emph{Positive document}\\ \textcolor{blue}{Tesla, Inc. (TSLA) reported Q1 FY 2022} earnings results\\ ... detailed revenues of \$\textbf{\textcolor{blue}{18.76 billion}} ...\\ \\ \emph{Negative document}\\ ...\textcolor{blue}{first-quarter earnings for 2022} ...\\ ...\textcolor{red}{Automotive revenue} reached \$\textcolor{red}{16.86 billion}...\end{tabular} \\ \midrule
\textbf{Responses} & \begin{tabular}[c]{@{}l@{}}Iga Swiatek defeated \textcolor{red}{Ons Jabeur} in the \textcolor{red}{second round}\\  of the Qatar Open 2022 to win the tournament.\end{tabular}                                                                                                                                                                                        & \begin{tabular}[c]{@{}l@{}}According to the document, the name of Apple's\\ headset is \textcolor{red}{Apple Reality Pro}.\end{tabular}                                                                                                                                                                                                & \begin{tabular}[c]{@{}l@{}}According to the financial results provided in the article,\\ Tesla's revenue in Q1 2022 was \$\textcolor{red}{16.86 billion}.\end{tabular}                                                                                                                                                                                                                               \\ \bottomrule
\end{tabular}}
\caption{Error cases of noise robustness, and only one positive document and one negative document are shown. The responses are generated by ChatGLM2-6B. The blue text indicates the matching parts between the document and the question or answer, while the red text highlights the non-matching parts.}
\label{tab:case1}
\end{table*}

Finally, we collect totally 600 base questions in RGB, and 200 additional questions for the information integration ability and 200 additional questions for counterfactual robustness ability. Half of the instances are in English, and the other half are in Chinese.

\subsection{Evaluation metrics}
The core of this benchmark is to evaluate whether LLMs can utilize the provided external documents to acquire knowledge and generate reasonable answers. We evaluate the responses of LLMs in order to measure above-mentioned four abilities of them. 

\textbf{Accuracy} is used to measure noise robustness and information integration. We employ an exact matching approach where if the generated text contains an exact match to the answer, it is considered as a correct answer.

\textbf{Rejection rate} is used to measure negative rejection. When only noisy documents are provided, LLMs should output the specific content -- ``I can not answer the question because of the insufficient information in documents.'' (We use instructions to inform the model.). If the model generates this content, it indicates a successful rejection. 

\textbf{Error detection rate} measures whether the model can detect the factual errors in the documents for counterfactual robustness. When the provided documents contain factual errors, the model should output the specific content -- ``There are factual errors in the provided documents.'' (We use instructions to inform the model.). If the model generates this content, it indicates that the model has detected erroneous information in the document.

\textbf{Error correction rate} measures whether the model can provide the correct answer after identifying errors for counterfactual robustness. The model is asked to generate the correct answer after identifying the factual errors. If the model generates the correct answer, it indicates that the model is capable of correcting errors in the document.



Considering that the model may not fully adhere to instructions, for rejection rate and error detection rate, we also use ChatGPT to conduct additional evaluation of the answers. Specifically, we  assess the model's responses by using instructions and demonstrations to determine if they can reflect information that is not present in the document or identify any factual errors.
\section{Experiments}
In this section, we evaluate the performance of various LLMs, analyze and discuss the results, summarizing the main challenges that existing LLMs encounter when using external knowledge.

\subsection{Settings}
\textbf{Task formats.} 
Due to contextual limitations, we provide 5 external documents for each question. In our experiments on noise robustness, we evaluate scenarios with noise ratios ranging from 0 to 0.8. To comprehensively evaluate the overall capabilities, we have adopted a unified instruction for each language, as shown in Figure~\ref{fig:io}. The experiments were conducted using an NVIDIA GeForce RTX 3090.

\textbf{Models}
We conduct evaluation on 6 state-of-the-art large language models which can generate both English and Chinese including ChatGPT~\citep{chatgpt}\footnote{We use gpt-3.5-turbo api in the experiments.}, ChatGLM-6B~\citep{chatglm}, ChatGLM2-6B~\citep{chatglm2}, Vicuna-7b-v1.3~\citep{vicuna2023}, Qwen-7B-Chat~\citep{qwen}, BELLE-7B-2M~\citep{BELLE}. 

\subsection{Results on Noise Robustness}
We evaluated the accuracy based on the different noise ratios in external documents, and the results are shown in Table~\ref{tab:result1}. We can see that:

\textbf{(1) RAG can effect improve the responses of LLMs.} LLMs have shown strong performance even in the presence of noise, indicating that RAG is a promising way for LLMs to generate accurate and reliable responses.

\textbf{(2) The increasing noise rate poses a challenge for RAG in LLMs.}  Specifically, when the noise ratio exceeds 80\%, the accuracy decreases significantly at a significance level of 0.05. For example, the performance of ChatGPT has decreased from 96.33\% to 76.00\%, while the performance of ChatGLM2-6B has decreased from 91.33\% to 57.33\%.

\subsubsection{Error Analysis.}
To better comprehend the negative impact of noise on model generation, we examined the incorrect answers and found that these errors typically originate from three reasons, as shown in Table~\ref{tab:case1}.

\textbf{(1) Long-distance information.} LLMs often face difficulty in identifying the correct answer from external documents when the information related to the question is distant from the information related to the answer. This scenario is quite common as longer texts are frequently encountered on the internet. In such cases, it is typical for the question's information to be initially presented at the start of the document and subsequently referred to using pronouns. In Table~\ref{tab:case1}, the question information (``Qatar Open 2022'') is only mentioned once at the beginning and is far from where the answer text ``Anett Kontaveit'' appears. This situation may cause LLMs to depend on information from other documents and create false impressions, i.e., hallucination.

\textbf{(2) Evidence uncertainty.} Before highly anticipated events, like the release of new Apple products or the announcement of the Oscars, there is often a significant amount of speculative information circulating on the internet. Although the relevant documents explicitly state that it is uncertain or speculative content, they can still impact on the retrieval-augmented generation of LLMs. In Table~\ref{tab:case1}, when the noise ratio increases, the content of erroneous documents is all about some people's predictions about the name of the headset (``Apple Reality Pro''). Even if there is a correct answer (``Vision Pro'') in the relevant documents, LLMs can still be misled by uncertain evidences.

\textbf{(3) Concept confusion.} The concepts in external documents may be similar to, but different from, the concepts in the question. This can cause confusion for LLMs and make LLMs generate incorrect answers. In Table~\ref{tab:case1}, the model answer focuses on the concept ``automotive revenue'' in the document rather than ``revenue'' in the question.

Based on the analysis above, we have identified certain limitations in LLMs regarding retrieval-augmented generation. To effectively handle the vast amount of noise present on the internet, further detailed enhancements are required for the model such as long documents modeling and precise concept comprehension.

\subsection{Results on Negative Rejection testbed}
We evaluated the rejection rate when only noise documents were provided. The results are shown in Table~\ref{tab:result2}. In addition to evaluating the rejection rate through exact matching (Rej in Table~\ref{tab:result2}), we also utilize ChatGPT to determine if the responses from the LLMs contain any rejection information (Rej$^*$ in Table~\ref{tab:result2}). We can see that: \textbf{Negative Rejection poses a challenge for RAG in LLMs.} The highest rejection rates for LLMs in English and Chinese were only 45\% and 43.33\%, respectively. This suggests that LLMs can be easily misled by noisy documents, leading to incorrect answers.

In addition, through comparing Rej and Rej$^*$, we found that LLMs fail to strictly follow instructions, and they often generate unpredictable responses, which make it hard to use them as state triggers (such as for recognizing rejection).

\begin{table}[]
\centering
\resizebox{0.3\textwidth}{!}{\begin{tabular}{@{}c|cc|cc@{}}
\toprule
Languages      & \multicolumn{2}{c|}{English}    & \multicolumn{2}{c}{Chinese}    \\ \midrule
               & Rej            & Rej$^*$        & Rej           & Rej$^*$        \\ \midrule
ChatGPT        & 24.67          & \textbf{45.00} & 5.33          & \textbf{43.33} \\
ChatGLM-6B     & 9.00           & 25.00          & 6.33          & 17.00          \\
ChatGLM2-6B    & 10.33          & 41.33          & 6.33          & 36.33          \\
Vicuna-7B-v1.3 & 17.00          & 33.33          & 3.37          & 24.67          \\
Qwen-7B-Chat   & \textbf{31.00} & 35.67          & \textbf{8.67} & 25.33          \\
BELLE-7B-2M    & 5.67           & 32.33          & 5.33          & 13.67          \\ \bottomrule
\end{tabular}}
\caption{The result of negative rejection. Rej means the rejection rate (\%) and Rej$^*$ means the rejection rate evaluated by ChatGPT. We can see that negative rejection poses a challenge for RAG in LLMs.}
\label{tab:result2}
\end{table}

\begin{table}[]
\centering
\resizebox{0.45\textwidth}{!}{\begin{tabular}{@{}l|l|l@{}}
\toprule
\multicolumn{1}{c|}{Question}                                                                               & \multicolumn{1}{c|}{Answer}                                      & \multicolumn{1}{c}{Response}                                                                                                                                                                                                                                                                             \\ \midrule
\begin{tabular}[c]{@{}l@{}}who will direct\\ Irredeemable film?\end{tabular}                                & \textbf{\begin{tabular}[c]{@{}l@{}}Jeymes\\ Samuel\end{tabular}} & \begin{tabular}[c]{@{}l@{}}The information provided in the document\\ is that Fox has attached ``The Big Short''\\ writer-director \textcolor{red}{Adam McKay} to its movie\\ adaptation of ``Irredeemable'' from Boom!\\ Comics. Therefore, the answer is Adam McKay.\end{tabular}                          \\ \midrule
\begin{tabular}[c]{@{}l@{}}Which country won the\\ most medals at the \\ 2022 Winter Olympics?\end{tabular} & \textbf{Norway}                                                  & \begin{tabular}[c]{@{}l@{}}Based on the information provided in the\\ document, the country that won the most medals\\ at the 2022 Winter Olympics is \textcolor{red}{Germany}.\\ Germany has won a total of 31 medals, including 14\\ gold medals, 14 bronze medals, and 31 silver medals.\end{tabular} \\ \bottomrule
\end{tabular}}
\caption{Error cases of negative rejection generated by ChatGLM2-6B. The red text highlights the error answers.}
\label{tab:case2}
\end{table}

We conduct case studies in Table~\ref{tab:case2}. The first error is because of \textbf{Evidence uncertainty}. Although the document only mentions contact with ``Adam McKay'' and does not explicitly state that he is the director of the movie, the model still concludes that he holds this role. The first error is because of \textbf{Concept confusion}. The information provided in the answer pertains to ``the 2018 Winter Olympics'' instead of ``the 2022 Olympics'' mentioned in the question.  
Retrieval-augmented generation poses a greater challenge of negative rejection compared to answer directly as it presents relevant documents that could potentially mislead the LLMs and result in incorrect responses. In future developments, it will be crucial for LLMs to enhance their ability to accurately match questions with the appropriate documents.

\subsection{Results on Information Integration testbed}
We evaluated the accuracy based on the different noise ratios in external documents, and the results are shown in Table~\ref{tab:result3}. When comparing the model to Table~\ref{tab:result1}, we observed that it has a weak information integration ability, which in turn affects its noise robustness. We can see that:

\textbf{(1) Information integration poses a challenge for RAG in LLMs.} Even without noise, the highest accuracy of LLMs can only reach 60\% and 67\% for English and Chinese, respectively. After adding noise, the highest accuracy decreases to 43\% and 55\%. These results suggest that LLMs struggle with integrating information effectively and are not well-suited for directly answering complex questions.

\textbf{(2) Complex questions are more challenging for RAG with noisy documents.} Performance decline becomes significant when the noise ratio is 0.4, but for simple problems, a significant decline occurs only at a noise ratio of 0.8 at a significance level of 0.05. This indicates that complex problems are more vulnerable to interference from noise. We speculate that this is because solving complex problems requires integrating information from multiple documents, and this information can be considered as noise to each other, making it harder for the model to extract relevant information from the documents.

\begin{table}[]
\centering
\resizebox{0.3\textwidth}{!}{\begin{tabular}{c|ccc|ccc}
\hline
      & \multicolumn{3}{c|}{English}            & \multicolumn{3}{c}{Chinese}             \\ \hline
Noise Ratio     & 0           & 0.2         & 0.4         & 0           & 0.2         & 0.4         \\ \hline
ChatGPT        & 55          & 51          & 34          & 63          & \textbf{58} & 47          \\
ChatGLM-6B     & 45          & 36          & 35          & 60          & 53          & 52          \\
ChatGLM2-6B    & 34          & 32          & 21          & 44          & 43          & 32          \\
Vicuna-7B-v1.3 & \textbf{60} & \textbf{53} & \textbf{43} & 43          & 36          & 25          \\
Qwen-7B-Chat   & 55          & 50          & 37          & \textbf{67} & 56          & \textbf{55} \\
BELLE-7B-2M    & 40          & 34          & 24          & 49          & 41          & 38          \\ \hline
\end{tabular}}
\caption{The experimental result of information integration measured by accuracy (\%) under different noise ratios. We can see that information integration poses a challenge for RAG in LLMs.}
\label{tab:result3}
\end{table}

\begin{table}[]
\centering
\resizebox{0.49\textwidth}{!}{\begin{tabular}{@{}l|l|l|c@{}}
\toprule
\multicolumn{1}{c|}{\textbf{Question}}                                                                        & \multicolumn{1}{c|}{\textbf{Answer}}                                                                                                                      & \multicolumn{1}{c|}{\textbf{Response}}                                                                                                                                 & Errors                                                                \\ \midrule
\begin{tabular}[c]{@{}l@{}}What groupings are France and\\ Germany in World Cup 2022?\end{tabular}            & \begin{tabular}[c]{@{}l@{}}\textcolor{blue}{Group D}\\ \textcolor{red}{Group E}\end{tabular}                                                              & \begin{tabular}[c]{@{}l@{}}France and Germany are in\\ \textcolor{blue}{Group D} of the 2022 FIFA\\ World Cup.\end{tabular}                                            & \textbf{\begin{tabular}[c]{@{}c@{}}Merging\\ Error\end{tabular}}      \\ \midrule
\begin{tabular}[c]{@{}l@{}}Who were the MVP of Super\\ Bowl 2022 and 2023?\end{tabular}                       & \begin{tabular}[c]{@{}l@{}}\textcolor{blue}{Cooper Kupp}\\ \textcolor{red}{Patrick}\\ \textcolor{red}{Mahomes}\end{tabular}                               & \begin{tabular}[c]{@{}l@{}}According to the provided\\ document, the MVP of Super\\ Bowl LVI was \textcolor{blue}{Cooper Kupp}\\ of the Los Angeles Rams.\end{tabular} & \textbf{\begin{tabular}[c]{@{}c@{}}Ignoring\\ Error\end{tabular}}     \\ \midrule
\begin{tabular}[c]{@{}l@{}}What films won the 2022 and\\ 2023 Academy Awards for\\ Best Picture?\end{tabular} & \begin{tabular}[c]{@{}l@{}}\textcolor{blue}{CODA}\\ \textcolor{red}{Everything}\\ \textcolor{red}{Everywhere}\\ \textcolor{red}{All at Once}\end{tabular} & \begin{tabular}[c]{@{}l@{}}The film \textcolor{blue}{CODA} won the\\ award for Best Picture at the\\ 95th Academy Awards\\ ceremony held on 2023.\end{tabular}         & \textbf{\begin{tabular}[c]{@{}c@{}}Misalignment\\ Error\end{tabular}} \\ \bottomrule
\end{tabular}}
\caption{Error cases of information integration, the responses are generated by ChatGLM2-6B. The blue and red texts represent the answers to two sub-questions.}
\label{tab:case3}
\end{table}

\subsubsection{Error Analysis.}

We conducted an error analysis on ChatGLM2-6B (noise ratio is 0). Apart from the similar errors founded in the noise robustness experiment (38\% of the total), there are also three types of unique errors. We have presented these cases in Table~\ref{tab:case3}. 

\textbf{(1) Merging Error (28\% of the total).} The model sometimes merges the answers of the two sub-questions, resulting in an error. It mistakenly uses the answer from one question to address both two questions. At this point, the model will disregard any documents related to one sub-question. For example, in Table~\ref{tab:case3}, it incorrectly states that Group D is the World Cup group for both France and Germany, while in fact Germany is actually assigned to Group E.

\textbf{(2) Ignoring Error (28\% of the total).} Sometimes, the model may ignore one of the sub-questions and only answer the other. This error occurs when the model lacks a complete understanding of the problem and fails to recognize that it consists of multiple sub-problems. As a result, the model only considers relevant documents for one sub-problem in order to generate an answer, disregarding the question posed by another sub-problem. For example, in Table~\ref{tab:case3}, the model only provides the answer for the MVP of Super Bowl 2022 and does not consider 2023.

\textbf{(3) Misalignment Error (6\% of the total).} Sometimes, the model incorrectly identifies the documents for one sub-question as the documents for another sub-question, leading to misaligned answers. For example, in Table~\ref{tab:case3}, the third answer has two errors: an ignoring error and a misalignment error. Firstly, the model only mentioned the Best Picture of the 2023 (95th) Academy Awards, completely disregarding the 2022 awards. Additionally, it incorrectly stated that ``CODA'' is the Best Picture of 2023 when it was actually awarded as the Best Picture in 2022.

The errors mentioned above are primarily caused by the limited understanding of complex questions, which hinders the ability to effectively utilize information from different sub-problems. The key lies in improving the model's reasoning capability. One possible solution is to use a chain-of-thought approach to break down complex problems~\citep{zhou2023leasttomost,xu2023searchinthechain,drozdov2023compositional}. However, these methods slow down the inference speed and cannot provide timely responses.

\begin{table}[]
\centering
\resizebox{0.33\textwidth}{!}{\begin{tabular}{@{}c|ccccc@{}}
\toprule
                & Acc & Acc$_{\text{doc}}$ & ED         & ED$^{*}$   & CR             \\ \midrule
ChatGPT-zh      & 91  & \textbf{17}        & 1          & 3          & 33.33          \\
Qwen-7B-Chat-zh & 77  & 12                 & 5          & 4          & 25.00          \\
ChatGPT-en      & 89  & 9                  & \textbf{8} & \textbf{7} & \textbf{57.14} \\ \bottomrule
\end{tabular}}
\caption{The result of counterfactual robustness. ACC is the accuracy (\%) of LLMs without external documents. ACC$_{\text{doc}}$ is the accuracy (\%) of LLMs with counterfactual documents. ED and ED$^{*}$ are error detection rates evaluated by exact matching and ChatGPT, respectively. CR is the error correction rate.}
\label{tab:result4}
\end{table}

\subsection{Results on Counterfactual Robustness testbed}
In order to ensure that LLMs possess relevant knowledge, we assess their performance by directly asking them questions. However, we found that most LLMs struggle to answer them correctly. To ensure a more reasonable evaluation, we only consider LLMs that have an accuracy rate of over 70\% as this threshold is relatively high and encompasses more LLMs. The results are shown in Table~\ref{tab:result4}. We present the following metrics: accuracy without any documents, accuracy with counterfactual documents, error detection rates, and error correction rates. We can see that It is hard for LLMs to identify and correct factual errors in the documents. This suggests that the model can be easily misled by documents containing incorrect facts.

It is important to note that retrieval-augmented generation is not designed to automatically address factual errors within a given context, as this contradicts the underlying assumption that the model lacks knowledge and relies on retrieved documents for additional information. However, this issue is crucial in practical applications due to the abundance of fake news on the internet. Existing LLMs do not have a safeguard to handle inaccurate responses caused by misinformation. In fact, they heavily depend on the information they retrieve. Even when LLMs contain the internal knowledge about the questions, they often trust false information that is retrieved. This presents significant a challenge for the future development of RAG in LLMs.

\section{Conclusion}
In this paper, we evaluated four abilities of retrieval-augmented generation in LLMs: noise robustness, negative rejection, information integration, and counterfactual robustness. To conduct the evaluation, we built Retrieval-Augmented Generation Benchmark (RGB). The instances of RGB are generated from latest news articles and the external documents obtained from search engines. The experimental results suggest that current LLMs have limitations in the 4 abilities. This indicates that there is still a significant amount of work needed to effectively apply RAG to LLMs. To ensure accurate and reliable responses from LLMs, it is crucial to exercise caution and carefully design for RAG.

\section{Acknowledgements}
This research work is supported by the National Natural Science Foundation of China under Grants no. 62122077, 62106251, 62306303, the CAS Project for Young Scientists in Basic Research under Grant No.YSBR-040.  Xianpei Han is sponsored by CCF- BaiChuan-Ebtech Foundation Model Fund.

\bibliography{aaai24}

\begin{thebibliography}{46}
\providecommand{\natexlab}[1]{#1}

\bibitem[{Adlakha et~al.(2023)Adlakha, BehnamGhader, Lu, Meade, and
  Reddy}]{adlakha2023evaluating}
Adlakha, V.; BehnamGhader, P.; Lu, X.~H.; Meade, N.; and Reddy, S. 2023.
\newblock Evaluating Correctness and Faithfulness of Instruction-Following
  Models for Question Answering.
\newblock arXiv:2307.16877.

\bibitem[{Bang et~al.(2023)Bang, Cahyawijaya, Lee, Dai, Su, Wilie, Lovenia, Ji,
  Yu, Chung, Do, Xu, and Fung}]{bang2023multitask}
Bang, Y.; Cahyawijaya, S.; Lee, N.; Dai, W.; Su, D.; Wilie, B.; Lovenia, H.;
  Ji, Z.; Yu, T.; Chung, W.; Do, Q.~V.; Xu, Y.; and Fung, P. 2023.
\newblock A Multitask, Multilingual, Multimodal Evaluation of ChatGPT on
  Reasoning, Hallucination, and Interactivity.
\newblock arXiv:2302.04023.

\bibitem[{Bian et~al.(2023)Bian, Liu, Han, Lin, Lu, He, and Sun}]{bian2023drop}
Bian, N.; Liu, P.; Han, X.; Lin, H.; Lu, Y.; He, B.; and Sun, L. 2023.
\newblock A Drop of Ink Makes a Million Think: The Spread of False Information
  in Large Language Models.
\newblock arXiv:2305.04812.

\bibitem[{Borgeaud et~al.(2022)Borgeaud, Mensch, Hoffmann, Cai, Rutherford,
  Millican, van~den Driessche, Lespiau, Damoc, Clark, de~Las~Casas, Guy,
  Menick, Ring, Hennigan, Huang, Maggiore, Jones, Cassirer, Brock, Paganini,
  Irving, Vinyals, Osindero, Simonyan, Rae, Elsen, and
  Sifre}]{borgeaud2022improving}
Borgeaud, S.; Mensch, A.; Hoffmann, J.; Cai, T.; Rutherford, E.; Millican, K.;
  van~den Driessche, G.; Lespiau, J.-B.; Damoc, B.; Clark, A.; de~Las~Casas,
  D.; Guy, A.; Menick, J.; Ring, R.; Hennigan, T.; Huang, S.; Maggiore, L.;
  Jones, C.; Cassirer, A.; Brock, A.; Paganini, M.; Irving, G.; Vinyals, O.;
  Osindero, S.; Simonyan, K.; Rae, J.~W.; Elsen, E.; and Sifre, L. 2022.
\newblock Improving language models by retrieving from trillions of tokens.
\newblock arXiv:2112.04426.

\bibitem[{Cai et~al.(2019{\natexlab{a}})Cai, Wang, Bi, Tu, Liu, Lam, and
  Shi}]{cai-etal-2019-skeleton}
Cai, D.; Wang, Y.; Bi, W.; Tu, Z.; Liu, X.; Lam, W.; and Shi, S.
  2019{\natexlab{a}}.
\newblock Skeleton-to-Response: Dialogue Generation Guided by Retrieval Memory.
\newblock In \emph{Proceedings of the 2019 Conference of the North {A}merican
  Chapter of the Association for Computational Linguistics: Human Language
  Technologies, Volume 1 (Long and Short Papers)}, 1219--1228. Minneapolis,
  Minnesota: Association for Computational Linguistics.

\bibitem[{Cai et~al.(2019{\natexlab{b}})Cai, Wang, Bi, Tu, Liu, and
  Shi}]{cai-etal-2019-retrieval}
Cai, D.; Wang, Y.; Bi, W.; Tu, Z.; Liu, X.; and Shi, S. 2019{\natexlab{b}}.
\newblock Retrieval-guided Dialogue Response Generation via a
  Matching-to-Generation Framework.
\newblock In \emph{Proceedings of the 2019 Conference on Empirical Methods in
  Natural Language Processing and the 9th International Joint Conference on
  Natural Language Processing (EMNLP-IJCNLP)}, 1866--1875. Hong Kong, China:
  Association for Computational Linguistics.

\bibitem[{Cao et~al.(2020)Cao, Dong, Wu, and Cheung}]{cao-etal-2020-factual}
Cao, M.; Dong, Y.; Wu, J.; and Cheung, J. C.~K. 2020.
\newblock Factual Error Correction for Abstractive Summarization Models.
\newblock In \emph{Proceedings of the 2020 Conference on Empirical Methods in
  Natural Language Processing (EMNLP)}, 6251--6258. Online: Association for
  Computational Linguistics.

\bibitem[{Chang et~al.(2023)Chang, Wang, Wang, Wu, Yang, Zhu, Chen, Yi, Wang,
  Wang, Ye, Zhang, Chang, Yu, Yang, and Xie}]{chang2023survey}
Chang, Y.; Wang, X.; Wang, J.; Wu, Y.; Yang, L.; Zhu, K.; Chen, H.; Yi, X.;
  Wang, C.; Wang, Y.; Ye, W.; Zhang, Y.; Chang, Y.; Yu, P.~S.; Yang, Q.; and
  Xie, X. 2023.
\newblock A Survey on Evaluation of Large Language Models.
\newblock arXiv:2307.03109.

\bibitem[{Chiang et~al.(2023)Chiang, Li, Lin, Sheng, Wu, Zhang, Zheng, Zhuang,
  Zhuang, Gonzalez, Stoica, and Xing}]{vicuna2023}
Chiang, W.-L.; Li, Z.; Lin, Z.; Sheng, Y.; Wu, Z.; Zhang, H.; Zheng, L.;
  Zhuang, S.; Zhuang, Y.; Gonzalez, J.~E.; Stoica, I.; and Xing, E.~P. 2023.
\newblock Vicuna: An Open-Source Chatbot Impressing GPT-4 with 90\%* ChatGPT
  Quality.

\bibitem[{Cui et~al.(2023)Cui, Li, Yan, Chen, and Yuan}]{cui2023chatlaw}
Cui, J.; Li, Z.; Yan, Y.; Chen, B.; and Yuan, L. 2023.
\newblock ChatLaw: Open-Source Legal Large Language Model with Integrated
  External Knowledge Bases.
\newblock arXiv:2306.16092.

\bibitem[{Drozdov et~al.(2023)Drozdov, Sch{\"a}rli, Aky{\"u}rek, Scales, Song,
  Chen, Bousquet, and Zhou}]{drozdov2023compositional}
Drozdov, A.; Sch{\"a}rli, N.; Aky{\"u}rek, E.; Scales, N.; Song, X.; Chen, X.;
  Bousquet, O.; and Zhou, D. 2023.
\newblock Compositional Semantic Parsing with Large Language Models.
\newblock In \emph{The Eleventh International Conference on Learning
  Representations}.

\bibitem[{Edward~Beeching(2023)}]{open-llm-leaderboard}
Edward~Beeching, N. H. S. H. N. L. N. R. O. S. L. T. T.~W.,
  Clémentine~Fourrier. 2023.
\newblock Open LLM Leaderboard.
\newblock
  \url{https://huggingface.co/spaces/HuggingFaceH4/open_llm_leaderboard}.

\bibitem[{Guo et~al.(2023)Guo, Zhang, Wang, Jiang, Nie, Ding, Yue, and
  Wu}]{guo2023close}
Guo, B.; Zhang, X.; Wang, Z.; Jiang, M.; Nie, J.; Ding, Y.; Yue, J.; and Wu, Y.
  2023.
\newblock How Close is ChatGPT to Human Experts? Comparison Corpus, Evaluation,
  and Detection.
\newblock arXiv:2301.07597.

\bibitem[{Guu et~al.(2020)Guu, Lee, Tung, Pasupat, and
  Chang}]{10.5555/3524938.3525306}
Guu, K.; Lee, K.; Tung, Z.; Pasupat, P.; and Chang, M.-W. 2020.
\newblock REALM: Retrieval-Augmented Language Model Pre-Training.
\newblock In \emph{Proceedings of the 37th International Conference on Machine
  Learning}, ICML'20. JMLR.org.

\bibitem[{He, Zhang, and Roth(2022)}]{he2022rethinking}
He, H.; Zhang, H.; and Roth, D. 2022.
\newblock Rethinking with Retrieval: Faithful Large Language Model Inference.
\newblock arXiv:2301.00303.

\bibitem[{Hendrycks et~al.(2021)Hendrycks, Burns, Basart, Zou, Mazeika, Song,
  and Steinhardt}]{hendrycks2021measuring}
Hendrycks, D.; Burns, C.; Basart, S.; Zou, A.; Mazeika, M.; Song, D.; and
  Steinhardt, J. 2021.
\newblock Measuring Massive Multitask Language Understanding.
\newblock In \emph{International Conference on Learning Representations}.

\bibitem[{Huang et~al.(2023)Huang, Bai, Zhu, Zhang, Zhang, Su, Liu, Lv, Zhang,
  Lei, Fu, Sun, and He}]{huang2023ceval}
Huang, Y.; Bai, Y.; Zhu, Z.; Zhang, J.; Zhang, J.; Su, T.; Liu, J.; Lv, C.;
  Zhang, Y.; Lei, J.; Fu, Y.; Sun, M.; and He, J. 2023.
\newblock C-Eval: A Multi-Level Multi-Discipline Chinese Evaluation Suite for
  Foundation Models.
\newblock \emph{arXiv preprint arXiv:2305.08322}.

\bibitem[{Izacard and Grave(2021)}]{izacard-grave-2021-leveraging}
Izacard, G.; and Grave, E. 2021.
\newblock Leveraging Passage Retrieval with Generative Models for Open Domain
  Question Answering.
\newblock In \emph{Proceedings of the 16th Conference of the European Chapter
  of the Association for Computational Linguistics: Main Volume}, 874--880.
  Online: Association for Computational Linguistics.

\bibitem[{Izacard et~al.(2022)Izacard, Lewis, Lomeli, Hosseini, Petroni,
  Schick, Dwivedi-Yu, Joulin, Riedel, and Grave}]{izacard2022atlas}
Izacard, G.; Lewis, P.; Lomeli, M.; Hosseini, L.; Petroni, F.; Schick, T.;
  Dwivedi-Yu, J.; Joulin, A.; Riedel, S.; and Grave, E. 2022.
\newblock Atlas: Few-shot Learning with Retrieval Augmented Language Models.
\newblock arXiv:2208.03299.

\bibitem[{Ji et~al.(2023)Ji, Lee, Frieske, Yu, Su, Xu, Ishii, Bang, Madotto,
  and Fung}]{10.1145/3571730}
Ji, Z.; Lee, N.; Frieske, R.; Yu, T.; Su, D.; Xu, Y.; Ishii, E.; Bang, Y.~J.;
  Madotto, A.; and Fung, P. 2023.
\newblock Survey of Hallucination in Natural Language Generation.
\newblock \emph{ACM Comput. Surv.}, 55(12).

\bibitem[{Lewis et~al.(2020)Lewis, Perez, Piktus, Petroni, Karpukhin, Goyal,
  K\"{u}ttler, Lewis, Yih, Rockt\"{a}schel, Riedel, and
  Kiela}]{10.5555/3495724.3496517}
Lewis, P.; Perez, E.; Piktus, A.; Petroni, F.; Karpukhin, V.; Goyal, N.;
  K\"{u}ttler, H.; Lewis, M.; Yih, W.-t.; Rockt\"{a}schel, T.; Riedel, S.; and
  Kiela, D. 2020.
\newblock Retrieval-Augmented Generation for Knowledge-Intensive NLP Tasks.
\newblock In \emph{Proceedings of the 34th International Conference on Neural
  Information Processing Systems}, NIPS'20. Red Hook, NY, USA: Curran
  Associates Inc.
\newblock ISBN 9781713829546.

\bibitem[{Li et~al.(2023{\natexlab{a}})Li, Rawat, Zaheer, Wang, Lukasik, Veit,
  Yu, and Kumar}]{li-etal-2023-large}
Li, D.; Rawat, A.~S.; Zaheer, M.; Wang, X.; Lukasik, M.; Veit, A.; Yu, F.; and
  Kumar, S. 2023{\natexlab{a}}.
\newblock Large Language Models with Controllable Working Memory.
\newblock In \emph{Findings of the Association for Computational Linguistics:
  ACL 2023}, 1774--1793. Toronto, Canada: Association for Computational
  Linguistics.

\bibitem[{Li et~al.(2023{\natexlab{b}})Li, Zhang, Dubois, Taori, Gulrajani,
  Guestrin, Liang, and Hashimoto}]{alpaca_eval}
Li, X.; Zhang, T.; Dubois, Y.; Taori, R.; Gulrajani, I.; Guestrin, C.; Liang,
  P.; and Hashimoto, T.~B. 2023{\natexlab{b}}.
\newblock AlpacaEval: An Automatic Evaluator of Instruction-following Models.
\newblock \url{https://github.com/tatsu-lab/alpaca_eval}.

\bibitem[{Li et~al.(2023{\natexlab{c}})Li, Zhu, Ma, Liu, and
  Shah}]{li2023chatgpt}
Li, X.; Zhu, X.; Ma, Z.; Liu, X.; and Shah, S. 2023{\natexlab{c}}.
\newblock Are ChatGPT and GPT-4 General-Purpose Solvers for Financial Text
  Analytics? An Examination on Several Typical Tasks.
\newblock arXiv:2305.05862.

\bibitem[{Liu, Zhang, and Liang(2023)}]{liu2023evaluating}
Liu, N.~F.; Zhang, T.; and Liang, P. 2023.
\newblock Evaluating Verifiability in Generative Search Engines.
\newblock arXiv:2304.09848.

\bibitem[{Maynez et~al.(2020)Maynez, Narayan, Bohnet, and
  McDonald}]{maynez-etal-2020-faithfulness}
Maynez, J.; Narayan, S.; Bohnet, B.; and McDonald, R. 2020.
\newblock On Faithfulness and Factuality in Abstractive Summarization.
\newblock In \emph{Proceedings of the 58th Annual Meeting of the Association
  for Computational Linguistics}, 1906--1919. Online: Association for
  Computational Linguistics.

\bibitem[{OpenAI(2022)}]{chatgpt}
OpenAI. 2022.
\newblock Chatgpt: Optimizing language models for dialogue.
\newblock \url{https://openai.com/blog/chatgpt}.

\bibitem[{Peng et~al.(2023)Peng, Galley, He, Cheng, Xie, Hu, Huang, Liden, Yu,
  Chen, and Gao}]{peng2023check}
Peng, B.; Galley, M.; He, P.; Cheng, H.; Xie, Y.; Hu, Y.; Huang, Q.; Liden, L.;
  Yu, Z.; Chen, W.; and Gao, J. 2023.
\newblock Check Your Facts and Try Again: Improving Large Language Models with
  External Knowledge and Automated Feedback.
\newblock arXiv:2302.12813.

\bibitem[{Qin et~al.(2023)Qin, Liang, Ye, Zhu, Yan, Lu, Lin, Cong, Tang, Qian,
  Zhao, Tian, Xie, Zhou, Gerstein, Li, Liu, and Sun}]{qin2023toolllm}
Qin, Y.; Liang, S.; Ye, Y.; Zhu, K.; Yan, L.; Lu, Y.; Lin, Y.; Cong, X.; Tang,
  X.; Qian, B.; Zhao, S.; Tian, R.; Xie, R.; Zhou, J.; Gerstein, M.; Li, D.;
  Liu, Z.; and Sun, M. 2023.
\newblock ToolLLM: Facilitating Large Language Models to Master 16000+
  Real-world APIs.
\newblock arXiv:2307.16789.

\bibitem[{QwenLM(2023)}]{qwen}
QwenLM. 2023.
\newblock Qwen-7B.
\newblock \url{https://github.com/QwenLM/Qwen-7B}.

\bibitem[{Raunak, Menezes, and
  Junczys-Dowmunt(2021)}]{raunak-etal-2021-curious}
Raunak, V.; Menezes, A.; and Junczys-Dowmunt, M. 2021.
\newblock The Curious Case of Hallucinations in Neural Machine Translation.
\newblock In \emph{Proceedings of the 2021 Conference of the North American
  Chapter of the Association for Computational Linguistics: Human Language
  Technologies}, 1172--1183. Online: Association for Computational Linguistics.

\bibitem[{Ren et~al.(2023)Ren, Wang, Qu, Zhao, Liu, Tian, Wu, Wen, and
  Wang}]{ren2023investigating}
Ren, R.; Wang, Y.; Qu, Y.; Zhao, W.~X.; Liu, J.; Tian, H.; Wu, H.; Wen, J.-R.;
  and Wang, H. 2023.
\newblock Investigating the Factual Knowledge Boundary of Large Language Models
  with Retrieval Augmentation.
\newblock arXiv:2307.11019.

\bibitem[{Shen et~al.(2023)Shen, Chen, Backes, and Zhang}]{shen2023chatgpt}
Shen, X.; Chen, Z.; Backes, M.; and Zhang, Y. 2023.
\newblock In ChatGPT We Trust? Measuring and Characterizing the Reliability of
  ChatGPT.
\newblock arXiv:2304.08979.

\bibitem[{Shi et~al.(2023)Shi, Min, Yasunaga, Seo, James, Lewis, Zettlemoyer,
  and tau Yih}]{shi2023replug}
Shi, W.; Min, S.; Yasunaga, M.; Seo, M.; James, R.; Lewis, M.; Zettlemoyer, L.;
  and tau Yih, W. 2023.
\newblock REPLUG: Retrieval-Augmented Black-Box Language Models.
\newblock arXiv:2301.12652.

\bibitem[{THUDM(2023{\natexlab{a}})}]{chatglm}
THUDM. 2023{\natexlab{a}}.
\newblock ChatGLM-6B.
\newblock \url{https://github.com/THUDM/ChatGLM-6B}.

\bibitem[{THUDM(2023{\natexlab{b}})}]{chatglm2}
THUDM. 2023{\natexlab{b}}.
\newblock ChatGLM2-6B.
\newblock \url{https://github.com/THUDM/ChatGLM2-6B}.

\bibitem[{Trivedi et~al.(2023)Trivedi, Balasubramanian, Khot, and
  Sabharwal}]{trivedi-etal-2023-interleaving}
Trivedi, H.; Balasubramanian, N.; Khot, T.; and Sabharwal, A. 2023.
\newblock Interleaving Retrieval with Chain-of-Thought Reasoning for
  Knowledge-Intensive Multi-Step Questions.
\newblock In \emph{Proceedings of the 61st Annual Meeting of the Association
  for Computational Linguistics (Volume 1: Long Papers)}, 10014--10037.
  Toronto, Canada: Association for Computational Linguistics.

\bibitem[{Wang et~al.(2019{\natexlab{a}})Wang, Pruksachatkun, Nangia, Singh,
  Michael, Hill, Levy, and Bowman}]{10.5555/3454287.3454581}
Wang, A.; Pruksachatkun, Y.; Nangia, N.; Singh, A.; Michael, J.; Hill, F.;
  Levy, O.; and Bowman, S.~R. 2019{\natexlab{a}}.
\newblock \emph{SuperGLUE: A Stickier Benchmark for General-Purpose Language
  Understanding Systems}.
\newblock Red Hook, NY, USA: Curran Associates Inc.

\bibitem[{Wang et~al.(2019{\natexlab{b}})Wang, Singh, Michael, Hill, Levy, and
  Bowman}]{wang2018glue}
Wang, A.; Singh, A.; Michael, J.; Hill, F.; Levy, O.; and Bowman, S.~R.
  2019{\natexlab{b}}.
\newblock {GLUE}: A Multi-Task Benchmark and Analysis Platform for Natural
  Language Understanding.
\newblock In \emph{International Conference on Learning Representations}.

\bibitem[{Xu et~al.(2023{\natexlab{a}})Xu, Liu, Yan, Xu, Si, Zhou, Yi, Gao,
  Sang, Zhang, Zhang, Peng, Huang, and Zhou}]{xu2023cvalues}
Xu, G.; Liu, J.; Yan, M.; Xu, H.; Si, J.; Zhou, Z.; Yi, P.; Gao, X.; Sang, J.;
  Zhang, R.; Zhang, J.; Peng, C.; Huang, F.; and Zhou, J. 2023{\natexlab{a}}.
\newblock CValues: Measuring the Values of Chinese Large Language Models from
  Safety to Responsibility.
\newblock arXiv:2307.09705.

\bibitem[{Xu et~al.(2023{\natexlab{b}})Xu, Pang, Shen, Cheng, and
  Chua}]{xu2023searchinthechain}
Xu, S.; Pang, L.; Shen, H.; Cheng, X.; and Chua, T.-S. 2023{\natexlab{b}}.
\newblock Search-in-the-Chain: Towards Accurate, Credible and Traceable Large
  Language Models for Knowledge-intensive Tasks.
\newblock arXiv:2304.14732.

\bibitem[{Yunjie~Ji(2023)}]{BELLE}
Yunjie~Ji, Y. G. Y. P. Q. N. B. M. X.~L., Yong~Deng. 2023.
\newblock BELLE: Bloom-Enhanced Large Language model Engine.
\newblock \url{https://github.com/LianjiaTech/BELLE}.

\bibitem[{Zhang et~al.(2023)Zhang, Aljunied, Gao, Chia, and
  Bing}]{zhang2023m3exam}
Zhang, W.; Aljunied, S.~M.; Gao, C.; Chia, Y.~K.; and Bing, L. 2023.
\newblock M3Exam: A Multilingual, Multimodal, Multilevel Benchmark for
  Examining Large Language Models.

\bibitem[{Zhong et~al.(2023)Zhong, Cui, Guo, Liang, Lu, Wang, Saied, Chen, and
  Duan}]{zhong2023agieval}
Zhong, W.; Cui, R.; Guo, Y.; Liang, Y.; Lu, S.; Wang, Y.; Saied, A.; Chen, W.;
  and Duan, N. 2023.
\newblock AGIEval: A Human-Centric Benchmark for Evaluating Foundation Models.
\newblock arXiv:2304.06364.

\bibitem[{Zhou et~al.(2023{\natexlab{a}})Zhou, Sch{\"a}rli, Hou, Wei, Scales,
  Wang, Schuurmans, Cui, Bousquet, Le, and Chi}]{zhou2023leasttomost}
Zhou, D.; Sch{\"a}rli, N.; Hou, L.; Wei, J.; Scales, N.; Wang, X.; Schuurmans,
  D.; Cui, C.; Bousquet, O.; Le, Q.~V.; and Chi, E.~H. 2023{\natexlab{a}}.
\newblock Least-to-Most Prompting Enables Complex Reasoning in Large Language
  Models.
\newblock In \emph{The Eleventh International Conference on Learning
  Representations}.

\bibitem[{Zhou et~al.(2023{\natexlab{b}})Zhou, Alon, Xu, Jiang, and
  Neubig}]{zhou2023docprompting}
Zhou, S.; Alon, U.; Xu, F.~F.; Jiang, Z.; and Neubig, G. 2023{\natexlab{b}}.
\newblock DocPrompting: Generating Code by Retrieving the Docs.
\newblock In \emph{The Eleventh International Conference on Learning
  Representations}.

\end{thebibliography}

\end{document}